\documentclass[11pt,a4paper]{article}
\usepackage[hyperref]{acl2021}
\usepackage{times}
\usepackage{latexsym}
\usepackage{amsmath}
\usepackage{multirow}
\usepackage{bbm}
\usepackage{graphicx}
\usepackage{amssymb}
\usepackage{booktabs}
\usepackage{adjustbox}
\usepackage{xspace}
\usepackage{algpseudocode}
\usepackage{algorithm}
\usepackage{times}
\usepackage{latexsym}
\usepackage{graphicx}
\usepackage{color, colortbl}
\usepackage{subcaption, makecell,cleveref}

\usepackage{microtype}

\usepackage{xspace,mfirstuc,tabulary}

\usepackage{array}
\newcommand{\PreserveBackslash}[1]{\let\temp=\\#1\let\\=\temp}
\newcolumntype{C}[1]{>{\PreserveBackslash\centering}p{#1}}
\newcolumntype{R}[1]{>{\PreserveBackslash\raggedleft}p{#1}}
\newcolumntype{L}[1]{>{\PreserveBackslash\raggedright}p{#1}}

\newcommand{\secref}[2][]{Section#1~\ref{sec:#2}}

\newcommand{\tabref}[2][]{Table#1~\ref{tab:#2}}
\newcommand{\figref}[2][]{Figure#1~\ref{fig:#2}}

\definecolor{LightCyan}{rgb}{0.88,1,1}
\aclfinalcopy 

\title{Evaluating the Efficacy of Summarization Evaluation across Languages}

\author{Fajri Koto \qquad Jey Han Lau \qquad Timothy Baldwin\\
	School of Computing and Information Systems \\
	The University of Melbourne \\
	\texttt{\small ffajri@student.unimelb.edu.au, jeyhan.lau@gmail.com, 
		tbaldwin@unimelb.edu.au} \\
}

\date{}

\begin{document}
\maketitle
\begin{abstract}

While automatic summarization evaluation methods developed for English 
are routinely applied to other languages, this is the first attempt to 
systematically quantify their panlinguistic efficacy. We take a 
summarization corpus for eight different languages, and manually 
annotate generated summaries for focus (precision) and coverage 
(recall).  Based on this, we evaluate 19 summarization evaluation 
metrics, and find that using multilingual BERT within BERTScore performs 
well across all languages, at a level above that for English.

\end{abstract}

\section{Introduction}

Although manual evaluation \cite{nenkova-passonneau-2004-evaluating,hardy-etal-2019-highres} of text summarization is more reliable and interpretable, most research on text summarization employs automatic evaluations such as ROUGE \cite{lin-2004-rouge}, METEOR \cite{lavie-agarwal-2007-meteor}, MoverScore \cite{zhao-etal-2019-moverscore}, and BERTScore \cite{zhang2020bertscore} because they are time- and cost-efficient.

In proposing these metrics, the authors measured correlation with human judgments based on English datasets that are not representative of modern summarization systems. For instance, \citet{lin-2004-rouge} use DUC\footnote{\url{https://duc.nist.gov/data.html}} 2001--2003 for ROUGE (meaning summaries were generated with largely outdated extractive summarization systems); \citet{zhao-etal-2019-moverscore} use the TAC\footnote{\url{https://tac.nist.gov/data/}} dataset for MoverScore (again, featuring summaries from largely defunct systems; see \citet{peyrard-2019-studying} and \citet{rankel-etal-2013-decade}); and \citet{zhang2020bertscore} developed BERTScore based on a machine translation corpus (WMT). 
In contemporaneous work, \citet{bhandari-etal-2020-evaluating} address 
this issue by annotating English CNN/DailyMail summaries produced by 
recent summarization models, and found disparities over results from 
TAC. 


Equally troublingly, ROUGE has become the default summarization evaluation metric for languages other than English \cite{hu-etal-2015-lcsts,scialom-etal-2020-mlsum,ladhak-etal-2020-wikilingua,koto-etal-2020-liputan6}, despite there being no systematic validation of its efficacy across other languages. The questions we ask in this study, therefore, are twofold: (1) \textit{How well do existing automatic metrics perform over languages other than English?} and (2) \textit{What automatic metric works best across different languages?}

\begin{figure}[t]
	\centering
	\includegraphics[width=\linewidth]{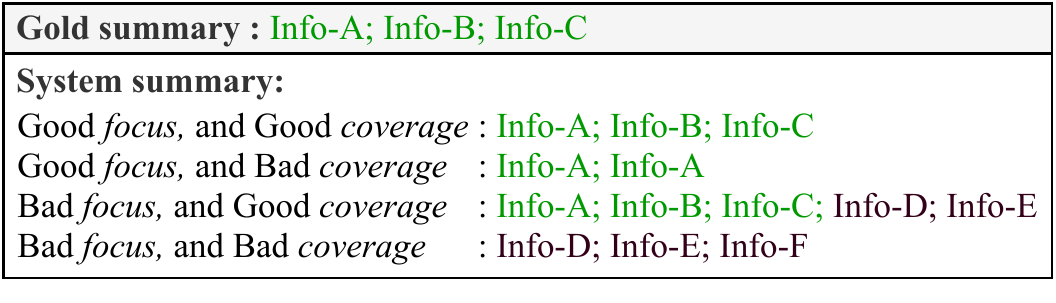}
	\caption{Illustration of focus and coverage.}
	\label{fig:definition}
\end{figure}

In this paper, we examine content-based summarization evaluation from the aspects of precision and recall, in the form of \textit{focus} and \textit{coverage} to compare system-generated summaries to ground-truth summaries (see \figref{definition}). As advocated by \citet{koto2020ffci}, focus and coverage are more interpretable and fine grained than the harmonic mean (F1 score) of ROUGE. This is also in line with the review of \citet{hardy-etal-2019-highres} on linguistic properties that have been manually evaluated in recent summarization research, who found precision and recall to be commonly used to complement ROUGE F1.


While it may seem more natural and reliable to evaluate focus and coverage based on the source document than the ground-truth summary, we use the ground-truth summary in this research for the following reasons. First, historically, validation of automatic evaluation metrics for summarization has been based primarily on ground-truth summaries (not source documents). Second, ROUGE \cite{lin-2004-rouge} was initially motivated and assessed based on coverage over the DUC datasets\footnote{DUC 2001, 2002, 2003} \cite{lin-hovy-2002-manual} using annotations based on reference summaries (not source documents). Third, although it is certainly possible to generate different summaries for the same source document, we argue that the variance in content is actually not that great, especially for single-document summarization.
Lastly, basing human evaluation (of focus and coverage) on the source article leads to more complicated annotation schemes, and has been shown to yield poor annotations \cite{nenkova-passonneau-2004-evaluating,fabbri2020summeval}.

In summary, this paper makes three contributions: (1) we carry out the first systematic attempt to quantify the efficacy of automatic summarization metrics over 8 linguistically-diverse languages, namely English (EN), Indonesian (ID), French (FR), Turkish (TR), Mandarin Chinese (ZH), Russian (RU), German (DE), and Spanish (ES); (2) we evaluate an extensive range of traditional and model-based metrics, and find BERTScore to be the best metric for evaluating both focus and coverage; and (3) we release a manually-annotated multilingual resource for summarization evaluation comprising 4,320 annotations. Data and code used in this paper is available at: \url{https://github.com/fajri91/Multi_SummEval}.

\section{Related Work}

As with much of NLP, research on automatic summarization metrics has been highly English-centric. \citet{graham-2015-evaluating} comprehensively evaluated 192 ROUGE variations based on the DUC-2004 (English) dataset. \citet{bhandari-etal-2020-evaluating} released a new (English) evaluation dataset by annotating CNN/DailyMail using simplified Pyramid \cite{nenkova-passonneau-2004-evaluating}. First, semantic content units (SCUs) were manually extracted from the reference, and crowd-workers were then asked to count the number of SCUs in the system summary. Their annotation procedure does not specifically consider focus, but is closely related to the coverage aspect of our work. Similarly, \citet{fabbri2020summeval} annotated the (English) CNN/DailyMail dataset for the four aspects of {coherence}, {consistency}, {fluency}, and {relevance}. While their work does not specifically study focus and coverage, relevance in their work can be interpreted as the harmonic mean of focus and coverage.

There is little work on summarization evaluation for languages other than English, and what work exists is primarily based on summaries generated by unsupervised extractive models dating back more than a decade, for a small handful of languages. Two years prior to ROUGE, \citet{saggion-etal-2002-meta} proposed a summarization metric using similarity measures for English and Chinese, based on cosine similarity, unit overlap, and the longest common subsequence (``LCS'') between reference and system summaries. In other work, \citet{saggion-etal-2010-multilingual} investigated {coverage}, {responsiveness}, and {pyramids} for several extractive models in English, French, and Spanish.

To the best of our knowledge, we are the first to systemically quantify the panlinguistic efficacy of evaluation metrics for modern summarization systems.

\section{Evaluation Metrics}

We assess a total of 19 different evaluation metrics that are commonly used in summarization research (noting that lesser-used metrics such as FRESA \cite{saggion-etal-2010-multilingual} and RESA \cite{cohan-goharian-2016-revisiting} are omitted from this study).

\textbf{ROUGE} \cite{lin-2004-rouge} measures the lexical overlap between the system and reference summary; based on the findings of \citet{graham-2015-evaluating}, we consider 7 variants in this paper: ROUGE-1 (unigram), ROUGE-2 (bigram), ROUGE-3 (trigram), ROUGE-L (LCS), ROUGE-S  (skip-bigram), ROUGE-SU (skip-bigram plus unigram), and ROUGE-W (weighted LCS).\footnote{\url{https://github.com/bheinzerling/pyrouge}}

\textbf{METEOR} \cite{lavie-agarwal-2007-meteor} performs word-to-word 
matching based on word-alignment, and was originally developed for MT 
but has recently been used for summarization evaluation 
\cite{see-etal-2017-get,chen-bansal-2018-fast,falke-gurevych-2019-fast,amplayo-lapata-2020-unsupervised}.\footnote{\url{http://www.cs.cmu.edu/~alavie/METEOR/}}

\textbf{BLEU} \cite{papineni-etal-2002-bleu} is a precision-based metric originally developed for MT, which measures the $n$-gram match between the reference and system summary. Based on the findings of \citet{graham-2015-evaluating}, we use BLEU-4 according to the SacreBLEU implementation \cite{post-2018-call}.\footnote{\url{https://github.com/mjpost/sacrebleu}}

\textbf{MoverScore} \cite{zhao-etal-2019-moverscore} measures the 
Euclidean distance between two contextualized BERT representations, and 
relies on soft alignments of words learned by solving an optimisation 
problem.\footnote{\url{https://github.com/AIPHES/emnlp19-moverscore}} We 
adapt use the default configuration 
(n-gram=1) over 5 different pre-trained models, as detailed below. Note 
that MoverScore is symmetric (i.e.\ MoverScore($x, y$) $=$ 
MoverScore($y, x$)), and as such is not designed to separately evaluate precision 
and recall.

\textbf{BERTScore} \cite{zhang2020bertscore} computes the similarity between BERT token embeddings of system and reference summaries based on soft overlap, in the form of precision, recall, and F1 scores.\footnote{ \url{https://github.com/Tiiiger/bert_score}} \citet{zhang2020bertscore} found that layer selection (i.e.\ which layer to source the token embeddings from) is critical to performance. Since layer selection in the original paper was based on MT datasets, we perform our own layer selection using a similar methodology as the authors, specifically considering precision and recall for focus and coverage, respectively.

For both MoverScore and BERTScore, we experiment with two classes of 
BERT-style model: (1) multilingual models, in the form of cased and 
uncased multilingual BERT \cite{devlin-etal-2019-bert}, and base and 
large XLM-R \cite{conneau-etal-2020-unsupervised}, for a total of 4 
models;\footnote{Note that both multilingual BERT and XLM were 
explicitly trained over all eight target languages used in this paper.} 
and (2) a monolingually-trained BERT for the given language, as listed 
in the Appendix. While we expect monolingual BERT models to perform 
better, we also focus on multilingual models, both to confirm whether 
this is the case, and to be able to draw findings for languages without 
monolingual models.

\section{Experimental Setup}

\begin{table}[t]
	\begin{center}
		\begin{adjustbox}{width=\linewidth}
			\begin{tabular}{ccccc}
				\toprule
				\multirow{3}{*}{\textbf{Lang}} & \multirow{3}{*}{\textbf{Quality (\%)}}  & \multicolumn{3}{c}{\bf Pearson correlation ($r$)} \\
				\cmidrule{3-5}
				&  &  \multicolumn{2}{c}{\bf Agreement} & {\bf Focus--} \\
				\cmidrule{3-4}
				& & \bf Focus & \bf Coverage & \bf Coverage\\
				\midrule
				EN & 90 & 0.47 & 0.46 & 0.58 \\
				ID & 97 & 0.64 & 0.63 & 0.80 \\
				FR & 98 & 0.63 & 0.65 & 0.71 \\
				TR & 97 & 0.74 & 0.79 & 0.74 \\
				ZH & 92 & 0.61 & 0.60 & 0.78 \\
				RU & 98 & 0.60 & 0.64 & 0.78 \\
				DE & 90 & 0.78 & 0.83 & 0.89 \\
				ES & 95 & 0.60 & 0.61 & 0.76 \\
				\bottomrule
			\end{tabular}
		\end{adjustbox}
	\end{center}
    \caption{Analysis of the annotations for each language, in terms of: 
    (1) average quality control score of approved HITs (\%); (2) one-vs-rest 
human agreement ($r$); and (3) correlation ($r$) between {focus} and {coverage}.}
	\label{tab:pearson}
\end{table}

For each language, we sample 135 documents from the test set of a pre-existing (single-document) summarization dataset: (1) CNN/DailyMail (English: \citet{hermann2015teaching}); (2) Liputan6 (Indonesian:  \citet{koto-etal-2020-liputan6}); (3) LCSTS (Chinese: \citet{hu-etal-2015-lcsts}); and (4) MLSUM (French, Turkish, Russian, German, Spanish: \citet{scialom-etal-2020-mlsum}). We source summaries based on two popular models: pointer generator network \cite{see-etal-2017-get} and BERT \cite{liu-lapata-2019-text,dong2019unified},\footnote{English, Indonesian and Chinese summaries were generated with the \citet{liu-lapata-2019-text} model, and the \citet{dong2019unified} model was used for the MLSUM-based languages.} and have 3 annotators annotate {focus} and {coverage} for each reference--system summary pair.\footnote{Summaries for all datasets except LCSTS were sourced from the authors of the dataset. For LCSTS, we trained the two models ourselves based on the training data.} The motivation for using BERT-based systems is that our study focuses on non-English summarization, where BERT-based models dominate.\footnote{BERT-based summaries are representative of transformer-based model, and the ROUGE score gap over state-of-the-art models \cite{zhang2020pegasus} for English is only $\sim$2 points.} The total number of resulting annotations is: 8 languages $\times$ 135 documents $\times$ 2 models $\times$ 2 criteria (= focus and coverage) $\times$ 3 annotators = 12,960.

\begin{table*}[t!]
	\footnotesize
	\begin{center}
		\begin{adjustbox}{width=1\linewidth}
			\begin{tabular}{lrrrrrrrrrrrrrrrrrrr}
				\toprule
				& \multicolumn{9}{c}{\textbf{Focus}} && \multicolumn{9}{c}{\textbf{Coverage}}\\
				\cmidrule{2-10}
				\cmidrule{12-20}
				& \textbf{EN} & \textbf{ID} & \textbf{FR} & \textbf{TR} & \textbf{ZH} & \textbf{RU} & \textbf{DE} & \textbf{ES} & \textbf{Avg} & & \textbf{EN} & \textbf{ID} & \textbf{FR} & \textbf{TR} & \textbf{ZH} & \textbf{RU} & \textbf{DE} & \textbf{ES} & \textbf{Avg} \\
				\midrule
				\multicolumn{20}{l}{\textbf{Traditional Metrics}} \\
				\midrule
				ROUGE-1 & 0.61 & 0.69 & 0.68 & 0.81 & 0.80 & 0.47 & 0.88 & 0.53 & 0.68 &  & 0.62 & 0.72 & 0.67 & 0.83 & 0.79 & 0.58 & 0.89 & 0.67 & 0.72 \\
				ROUGE-2 & 0.57 & 0.63 & 0.67 & 0.80 & 0.76 & 0.48 & 0.87 & 0.61 & 0.67 &  & 0.56 & 0.66 & 0.71 & 0.79 & 0.75 & 0.59 & 0.89 & 0.67 & 0.70 \\
				ROUGE-3 & 0.46 & 0.53 & 0.59 & 0.76 & 0.67 & 0.31 & 0.85 & 0.54 & 0.59 &  & 0.48 & 0.57 & 0.63 & 0.74 & 0.66 & 0.46 & 0.88 & 0.58 & 0.62 \\
				ROUGE-L & 0.60 & 0.69 & 0.68 & 0.81 & 0.79 & 0.46 & 0.87 & 0.54 & 0.68 &  & 0.61 & 0.72 & 0.67 & 0.83 & 0.79 & 0.59 & 0.89 & 0.67 & 0.72 \\
				ROUGE-S & 0.59 & 0.65 & 0.60 & 0.78 & 0.70 & 0.46 & 0.85 & 0.51 & 0.64 &  & 0.60 & 0.69 & 0.67 & 0.78 & 0.73 & 0.53 & 0.89 & 0.64 & 0.69 \\
				ROUGE-SU & 0.59 & 0.66 & 0.61 & 0.78 & 0.72 & 0.43 & 0.85 & 0.50 & 0.64 &  & 0.60 & 0.70 & 0.68 & 0.78 & 0.75 & 0.56 & 0.89 & 0.65 & 0.70 \\
				ROUGE-W.12 & 0.60 & 0.67 & 0.67 & 0.81 & 0.78 & 0.44 & 0.87 & 0.53 & 0.67 &  & 0.58 & 0.69 & 0.67 & 0.81 & 0.78 & 0.59 & 0.89 & 0.66 & 0.71 \\
				METEOR & 0.47 & 0.67 & 0.64 & 0.74 & 0.81 & 0.55 & 0.83 & 0.60 & 0.66 &  & 0.63 & 0.71 & 0.64 & 0.80 & 0.78 & 0.58 & 0.89 & 0.69 & 0.72 \\
				BLEU-4 & 0.46 & 0.56 & 0.64 & 0.70 & 0.70 & 0.39 & 0.85 & 0.50 & 0.60 &  & 0.48 & 0.58 & 0.59 & 0.67 & 0.69 & 0.31 & 0.85 & 0.54 & 0.59 \\
				
				\midrule
				\multicolumn{20}{l}{\textbf{MoverScore}} \\
				\midrule
				mono-BERT & 0.58 & 0.65 & 0.71 & 0.82 & 0.77 & 0.49 & 0.89 & 0.59 & 0.69 &  & 0.59 & 0.62 & 0.67 & 0.78 & 0.77 & 0.41 & 0.88 & 0.61 & 0.67 \\
				mBERT (cased)  & 0.54 & 0.68 & 0.77 & 0.79 & 0.76 & \bf 0.60 & 0.88 & 0.63 & 0.70 &  & 0.52 & 0.69 & 0.72 & 0.75 & 0.75 & 0.49 & 0.85 & 0.68 & 0.68 \\
				mBERT (uncased)  & 0.59 & 0.69 & \bf 0.78 & 0.81 & 0.76 & \bf 0.60 & 0.89 & \bf 0.67 & \bf 0.72 &  & 0.59 & 0.69 & 0.75 & 0.77 & 0.75 & 0.50 & 0.86 & 0.70 & 0.70 \\
				XLM (base) & 0.53 & 0.64 & 0.69 & 0.80 & 0.71 & 0.35 & 0.87 & 0.56 & 0.64 &  & 0.58 & 0.62 & 0.63 & 0.74 & 0.69 & 0.22 & 0.85 & 0.64 & 0.62 \\
				XLM (large) & 0.51 & 0.58 & 0.68 & 0.79 & 0.57 & 0.33 & 0.87 & 0.53 & 0.61 &  & 0.55 & 0.62 & 0.59 & 0.72 & 0.58 & 0.21 & 0.84 & 0.56 & 0.58 \\
				
				\midrule
				\multicolumn{20}{l}{\textbf{BERTScore}} \\
				\midrule
				mono-BERT & \bf 0.62 & \bf 0.71 & 0.73 & \bf 0.83 & \bf 0.82 & 0.51 & \bf 0.91 & \bf 0.67 & \bf 0.72 &  & 0.66 & \bf 0.74 & \bf 0.77 & \bf 0.88 & \bf 0.80 & 0.65 & \bf 0.92 & \bf 0.74 & \bf 0.77 \\
				mBERT (cased)  & 0.56 & \bf 0.71 & 0.73 & \bf 0.83 & 0.78 & 0.56 & 0.90 & 0.59 & 0.71 &  & \bf 0.67 & 0.73 & 0.70 & 0.87 & 0.79 & \bf 0.72 & 0.90 & 0.71 & 0.76 \\
				mBERT (uncased)  & 0.61 & \bf 0.71 & 0.72 & \bf 0.83 & 0.79 & 0.55 & 0.90 & 0.62 & \bf 0.72 &  & 0.64 & \bf 0.74 & 0.72 & 0.87 & 0.79 & 0.70 & 0.90 & 0.71 & 0.76 \\
				XLM (base) & 0.59 & 0.65 & 0.67 & \bf 0.83 & 0.79 & 0.34 & 0.89 & 0.58 & 0.67 &  & 0.64 & 0.71 & 0.66 & 0.86 & 0.73 & 0.67 & 0.90 & 0.70 & 0.74 \\
				XLM (large) & 0.60 & 0.66 & 0.68 & \bf 0.83 & 0.79 & 0.42 & 0.90 & 0.60 & 0.69 &  & 0.65 & 0.70 & 0.69 & 0.86 & 0.74 & 0.66 & 0.90 & 0.70 & 0.74 \\
				
				\midrule 
				\bf Human performance& 0.47 & 0.64 & 0.63 & 0.74 & 0.61 & 0.60 & 0.78 & 0.60 & 0.63 & & 0.46 & 0.63 & 0.65 & 0.79 & 0.60 & 0.64 & 0.83 & 0.61 & 0.65 \\ 
				\bottomrule
			\end{tabular}
		\end{adjustbox}
	\end{center}
	\caption{Pearson correlation ($r$) between automatic metrics and human 
		judgments (for Pointer Generator and BERT models combined). We compute the precision and recall of ROUGE and BERTScore for focus and coverage, respectively. BERTScore uses the optimized layer, and other metrics are computed using the default configuration of the original implementation.}
	\label{tab:result}
\end{table*}

\begin{table}[t]
	\scriptsize
	\begin{center}
		\begin{adjustbox}{width=0.8\linewidth}
			\begin{tabular}{lrr}
				\toprule
				\multirow{2}{*}{\textbf{Model}} & \multicolumn{2}{c}{\bf Universal layer} \\
				\cmidrule{2-3}
				& \bf Focus & \bf Coverage \\
				\midrule
				mBERT (cased) & 12 & 5 \\
				mBERT (uncased) & 12 & 6 \\
				XLM-R (base) & 4 & 4 \\ 
				XLM-R (large) & 10 & 9 \\
				\bottomrule
			\end{tabular}
		\end{adjustbox}
	\end{center}
	\caption{Recommended layers for multilingual models.}
	\label{tab:universal_layer}
\end{table}

For annotation, we used Amazon Mechanical Turk\footnote{\url{https://www.mturk.com}} with
the customized Direct Assessment (``DA'') method \cite{graham-etal-2015-accurate, graham2017can}, which has become the de facto for MT evaluation in WMT.
For each HIT (100 samples), DA incorporates 10 pre-annotated samples for quality control. Crowd-sourced workers are given two texts and asked the question (in the local language): \textit{How much information contained in the second text can also be found in the first text?} We combine {focus} and {coverage} annotation into 1 task, as the only thing that differentiates them is the ordering of the system and reference summaries, which is opaque to the annotators.\footnote{For focus, the first text is the reference and the second text the system summary; for coverage, the order is reversed.} Workers responded by scoring based via a slider button (continuous scale of 1--100).\footnote{See Appendix for the MTurk annotation interface for each language.}

For each HIT, we create 10 samples for quality control: 5 samples are random pairs (should be scored 0) and the remaining 5 samples are repetitions of the same summary with minor edits (should be scored 100).
For each language, we asked a native speaker to translate all 
instructions and the annotation interface. For a single HIT, we paid 
USD\$13, and set the HIT approval rate to 95\%. For HITs 
to be included in the annotated data, a quality 
control score of at least 7 out of 10 needed to be achieved. HITs below 
this threshold were re-run (ensuring they were not completed by a worker 
who had already completed that HIT), until three above-threshold 
annotations were obtained.\footnote{We approved all HITs with at least 30 minutes working time and a  minimum quality control score of 5, irrespective of whether they passed the higher quality-control threshold required for the ground truth.} 
For each language, the HIT approval rate is set to 95\% (with the number of HITs approved varying across languages). The annotation for English was restricted to US-based workers, and for other languages except Chinese was based on countries where the language is an official language.\footnote{In MTurk, we did not set a specific location for Chinese because we found there are no workers in China.}

To obtain focus and coverage values, we follow standard practice in DA in $z$-scoring the scores from each annotator, and then averaging. 

\section{Results}

\subsection{Annotation Results}
\label{sec:anno}

In \tabref{pearson}, we present the results of the human annotation. We 
first normalize the ratings from each HIT into a $z$-score, and 
one-vs-rest Pearson correlation (excluding quality control items) to 
provide an estimate of human agreement/performance.\footnote{We follow 
\citet{lau-etal-2020-furiously} in computing one-vs-rest correlation: we randomly 
isolate a worker's score (for each sample) and compare it against the 
mean score of the rest using Pearson's $r$, and repeat this for 
1000 trials to get the mean correlation.}  For all languages, we observe 
     that the average quality and human agreement is moderately high.  
However, the agreement does vary, and it affects the interpretation of 
the correlation scores when we assess the automatic metrics later.  Note 
also that we get the lowest score for English, meaning the results for 
non-English languages are actually more robust.\footnote{The relative 
quality for different languages largely coincides with the findings of 
\citet{pavlick-etal-2014-language}.}

Although focus and coverage are positively correlated in 
\tabref{pearson}, the distribution of scores varies quite a bit 
between languages: English annotation variance is higher than the other languages, and 
has the lowest correlation between focus and coverage ($r$ = 0.57); for 
French, Russian, and Spanish, summaries generally have low focus and 
coverage (for more details, see scatterplots of focus-coverage in 
\figref{foc_cov} of the Appendix). 


\subsection{Correlation with Automatic Evaluation}

In \tabref{result} we present the Pearson correlation between the human 
annotations and various automatic metrics, broken down across language 
and focus vs.\ coverage, and (naively) aggregated across languages in the form of the average 
correlation. We also include the one-vs-rest annotator correlation 
(\secref{anno}) in the last row, as it can be interpreted as the average 
performance of a single annotator. Recognizing the sensitivity of Pearson's correlation to outliers \cite{mathur-etal-2020-tangled}, we manually examined the distribution of scores for all language--system combinations for outliers (and present all scatterplots in \figref{foc_cov} of the Appendix).

The general pattern is consistent across languages: BERTScore performs better than other metrics in terms of both focus and coverage. This finding is consistent with that of \citet{fabbri2020summeval} wrt expert annotations of relevance (interpreted as the harmonic mean of our focus and coverage). ROUGE-1 and ROUGE-L are overall the best versions of ROUGE, while BLEU-4 performs the worst. For coverage, METEOR tends to be competitive with ROUGE-1, especially for EN, FR, DE, and ES, in large part because these languages are supported by the METEOR lemmatization package.

For some pre-trained models, MoverScore is competitive with BERTScore, although the average correlation is lower, especially for coverage. 

We perform layer selection for BERTScore by selecting the layer that produces the highest correlation. For monolingual BERT the selection is based on the average correlation across the two summarization models, while for the multilingual models it is based on overall result across the 8 languages $\times$ 2 models. \tabref{universal_layer} details the recommended layer for computing BERTScore for each of the multilingual models.\footnote{Recommended layers for monolingual BERT are detailed in the Appendix.} 

We observe that BERTScore with monolingual BERT performs the best, at an average of 0.72 and 0.77 for focus and coverage, resp., but only marginally above the best of the multilingual models, namely mBERT uncased (0.72 and 0.76, resp.). Given that layer selection here was performed universally across all languages (to ensure generalizability to other languages), our overall recommendation for the best metric to use is BERTScore with mBERT uncased.

When we compare the metric results to the one-vs-rest single-annotator
performance from \Cref{tab:pearson}, we see a positive correspondence
between the relative scores for annotator agreement and metric
performance, which we suspect is largely an artefact of data quality
(i.e.\ the metrics are assessed to perform better for languages with
high agreement because the quality of the ground-truth is higher), but
further research is required to confirm this.  Generally the best
metrics tend to outperform single-annotator performance substantially
($>$0.10), suggesting these metrics are more reliable than a single
annotator.


\section{Conclusion}
In this work, we developed a novel dataset for assessing automatic evaluation metrics for focus and coverage across a broad range of languages and datasets. We found that BERTScore is the best metric for the vast majority of languages, and advocate that this metric be used for summarization evaluation across different languages in the future, supplanting ROUGE.


\section*{Acknowledgments}
We are grateful to the anonymous reviewers for
their helpful feedback and suggestions.
The first author is
supported by the Australia Awards Scholarship (AAS), funded by the Department of Foreign Affairs and Trade (DFAT), Australia.
This research was undertaken using the LIEF HPC-GPGPU Facility hosted at The University of Melbourne. This facility was established with the assistance of LIEF Grant LE170100200.

\bibliographystyle{acl_natbib}
\bibliography{anthology,acl2021}

\clearpage
\onecolumn
\appendix

\section{Supplementary Materials}
\label{sec:supplementary}

\begin{figure*}[ht]
	\centering
	\includegraphics[width=\textwidth]{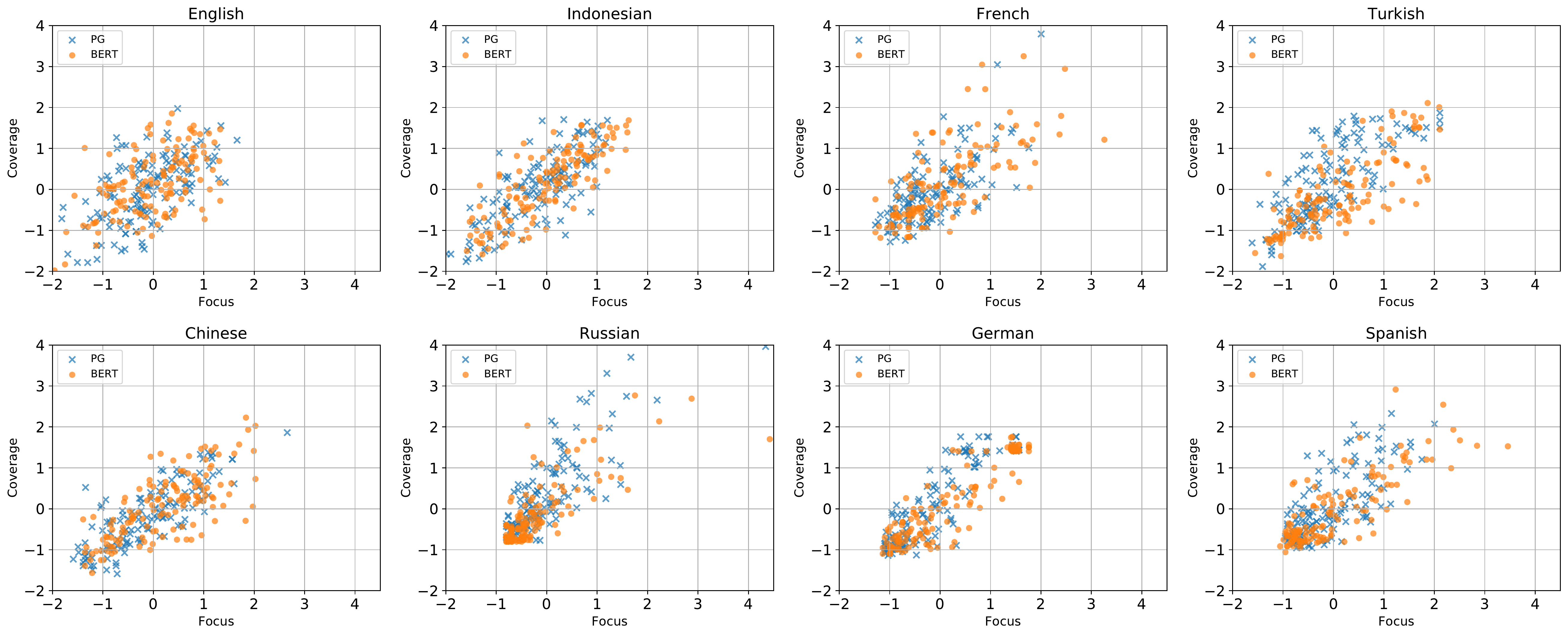}
	\caption{Annotation result (focus vs.\ coverage) after $z$-score normalization for each of the 8 languages.}
	\label{fig:foc_cov}
\end{figure*}

\begin{table*}[ht]
    \scriptsize
	\begin{center}
		\begin{adjustbox}{width=0.85\linewidth}
			\begin{tabular}{llcc}
				\toprule
				\multirow{2}{*}{\textbf{Lang}} &\multirow{2}{*}{\textbf{Model}} & \multicolumn{2}{c}{\bf Recommended layer} \\
				\cmidrule{3-4}
				& & \bf Focus & \bf Coverage \\
				\midrule
				EN & \texttt{bert-base-uncased} \cite{devlin-etal-2019-bert} & 1 & 2\\
				ID & \texttt{indolem/indobert-base-uncased} \cite{koto-etal-2020-indolem} & 2 & 2\\
				ZH & \texttt{bert-base-chinese} \cite{devlin-etal-2019-bert} & 8 & 9\\
				FR & \texttt{camembert-base} \cite{martin-etal-2020-camembert} & 10 & 9\\
				TR & \texttt{dbmdz/bert-base-turkish-uncased} & 12 & 4\\
				RU & \texttt{DeepPavlov/rubert-base-cased} \cite{kuratov2019adaptation} & 4 & 12\\
				DE & \texttt{bert-base-german-dbmdz-uncased} & 12 & 12\\
				ES & \texttt{dccuchile/bert-base-spanish-wwm-uncased} & 4 & 4\\
				
				\bottomrule
			\end{tabular}
		\end{adjustbox}
	\end{center}
	\caption{Recommended layers for computing focus and coverage via BERTScore with monolingual model.}
	\label{tab:other_layer}
\end{table*}

\begin{table*}[ht]
	\footnotesize
	\begin{center}
		\begin{adjustbox}{width=0.85\linewidth}
			\begin{tabular}{lccR{1.2cm}R{1cm}R{1cm}R{0.7cm}R{0.7cm}R{0.7cm}R{0.7cm}R{0.7cm}R{0.7cm}}
				\toprule
				 \multirow{2}{*}{\bf Language} & \multirow{2}{*} {\bf ISO} & \multirow{2}{*}{\bf Data } & \multicolumn{3}{c}{\bf Data Split} &  \multicolumn{3}{c}{\bf Pointer Generator} & \multicolumn{3}{c}{\bf BERT}
				 \\
				 & & & \bf Train & \bf Dev & \bf Test & \bf R1 & \bf R2 & \bf RL & \bf R1  & \bf R2 & \bf RL \\
				\midrule
				English & EN & CNN/DailyMail & 287,226 & 13,368 & 11,490 & 39.53 & 17.28 & 36.38 & 42.13 & 19.60 & 39.18\\
				Indonesian & ID & Liputan6 & 193,883 & 10,972 & 10,792 & 36.10 & 19.19 &	33.56 & 41.08 & 22.85 & 38.01\\
				Chinese & ZH & LCSTS & 2,400,591 & 8,672 & 725 & 32.39 & 19.92 & 29.45 & 38.47 & 25.45 & 35.30 \\
				French & FR & MLSUM & 392,902 & 16,059 & 15,828 & 26.50 & 9.49 & 20.30 & 28.52 & 11.73 & 22.51 \\
				Turkish & TR & MLSUM & 249,277 & 11,565 & 12,775 & 39.77 & 26.45 & 36.12 & 41.28 & 28.16 & 37.79 \\
				Russian & RU & MLSUM & 25,556 & 750 & 757 & 5.39 & 0.60 & 4.62 & 6.01 & 1.02 & 5.75 \\
				German & DE & MLSUM & 220,887 & 11,394 & 10,701 & 36.86 & 27.06 & 35.04 & 44.11 & 33.99 & 42.10 \\
				Spanish & ES & MLSUM & 266,367 & 10,358 & 13,920 & 25.05 & 7.44 & 19.53 & 26.48 & 9.59 & 21.69 \\
				\bottomrule
			\end{tabular}
		\end{adjustbox}
	\end{center}
	\caption{Details of datasets and ROUGE scores of summarization models used in this study. Other than for Chinese, we use summaries provided by the respective authors. For MLSUM, we report slightly different ROUGE-L scores because we use the original ROUGE package.}
	\label{tab:data}
\end{table*}

\begin{table*}[t!]
	\footnotesize
	\begin{center}
		\begin{adjustbox}{width=1\linewidth}
			\begin{tabular}{lrrrrrrrrrrrrrrrrrrr}
				\toprule
				\multirow{2}{*}{\textbf{Metrics}} & \multicolumn{9}{c}{\textbf{POINTER GENERATOR}} && \multicolumn{9}{c}{\textbf{BERT}}\\
				\cmidrule{2-10}
				\cmidrule{12-20}
				& \textbf{EN} & \textbf{ID} & \textbf{FR} & \textbf{TR} & \textbf{ZH} & \textbf{RU} & \textbf{DE} & \textbf{ES} & \textbf{Avg} & & \textbf{EN} & \textbf{ID} & \textbf{FR} & \textbf{TR} & \textbf{ZH} & \textbf{RU} & \textbf{DE} & \textbf{ES} & \textbf{Avg} \\
				\midrule
				\multicolumn{20}{l}{\textbf{Traditional Metrics}} \\
				\midrule
				ROUGE-1 & 0.59 & 0.68 & 0.54 & 0.82 & 0.81 & 0.52 & 0.85 & 0.50 & 0.66 &  & 0.61 & 0.70 & 0.73 & \bf 0.81 & 0.78 & 0.59 & 0.89 & 0.54 & 0.71 \\
				ROUGE-2 & 0.60 & 0.59 & 0.56 & 0.83 & 0.78 & 0.54 & 0.86 & 0.61 & 0.67 &  & 0.53 & 0.65 & 0.71 & 0.77 & 0.73 & 0.56 & 0.87 & 0.61 & 0.68 \\
				ROUGE-3 & 0.49 & 0.52 & 0.49 & 0.80 & 0.68 & 0.39 & 0.85 & 0.56 & 0.60 &  & 0.43 & 0.53 & 0.63 & 0.73 & 0.63 & 0.29 & 0.85 & 0.54 & 0.58 \\
				ROUGE-L & 0.59 & \bf 0.69 & 0.55 & 0.83 & 0.81 & 0.51 & 0.85 & 0.50 & 0.67 &  & 0.60 & 0.68 & 0.73 & 0.80 & 0.75 & 0.58 & 0.88 & 0.57 & 0.70 \\
				ROUGE-S & 0.60 & 0.62 & 0.48 & 0.79 & 0.70 & 0.55 & 0.83 & 0.50 & 0.63 &  & 0.58 & 0.67 & 0.64 & 0.76 & 0.69 & 0.59 & 0.86 & 0.51 & 0.66 \\
				ROUGE-SU & 0.59 & 0.63 & 0.50 & 0.79 & 0.72 & 0.55 & 0.83 & 0.50 & 0.64 &  & 0.58 & 0.68 & 0.66 & 0.77 & 0.70 & 0.60 & 0.86 & 0.50 & 0.67 \\
				ROUGE-W.12 & 0.60 & 0.67 & 0.56 & 0.83 & 0.81 & 0.52 & 0.86 & 0.49 & 0.66 &  & 0.60 & 0.66 & 0.72 & 0.79 & 0.74 & 0.55 & 0.88 & 0.57 & 0.69 \\
				METEOR & 0.49 & 0.65 & 0.51 & 0.82 & 0.85 & 0.52 & 0.86 & 0.68 & 0.67 &  & 0.45 & 0.68 & 0.70 & 0.71 & 0.77 & 0.52 & 0.85 & 0.58 & 0.66 \\
				BLEU-4 & 0.51 & 0.57 & 0.60 & 0.78 & 0.75 & 0.46 & 0.86 & 0.59 & 0.64 &  & 0.43 & 0.54 & 0.66 & 0.64 & 0.65 & 0.51 & 0.85 & 0.46 & 0.59 \\
				
				\midrule
				\multicolumn{20}{l}{\textbf{MoverScore}} \\
				\midrule
				mono-BERT & 0.62 & 0.67 & 0.63 & \bf 0.88 & 0.80 & \bf 0.65 & \bf 0.90 & 0.60 & 0.72 &  & 0.54 & 0.63 & 0.74 & 0.77 & 0.73 & 0.61 & 0.89 & 0.58 & 0.69 \\
				mBERT (cased)  & 0.57 & 0.62 & 0.71 & 0.84 & 0.79 & 0.60 & 0.88 & 0.71 & 0.72 &  & 0.51 & 0.71 & \bf 0.78 & 0.75 & 0.73 & 0.68 & 0.88 & 0.59 & 0.70 \\
				mBERT (uncased)  & 0.63 & 0.65 & \bf 0.76 & \bf 0.88 & 0.79 & 0.57 & 0.89 & \bf 0.74 & \bf 0.74 &  & 0.54 & 0.71 & \bf 0.78 & 0.76 & 0.73 & 0.68 & 0.88 & 0.63 & 0.71 \\
				XLM (base) & 0.56 & 0.61 & 0.60 & 0.86 & 0.73 & 0.32 & 0.86 & 0.71 & 0.66 &  & 0.49 & 0.63 & 0.71 & 0.77 & 0.68 & 0.52 & 0.89 & 0.51 & 0.65 \\
				XLM (large) & 0.53 & 0.57 & 0.60 & 0.84 & 0.62 & 0.30 & 0.87 & 0.63 & 0.62 &  & 0.48 & 0.58 & 0.69 & 0.76 & 0.52 & 0.42 & 0.87 & 0.46 & 0.60 \\
				
				\midrule
				\multicolumn{20}{l}{\textbf{BERTScore}} \\
				\midrule
				mono-BERT & 0.62 & 0.68 & 0.71 & 0.86 & 0.82 & 0.42 & \bf 0.90 & 0.69 & 0.71 &  & 0.62 & 0.73 & 0.72 & 0.80 & \bf 0.81 & \bf 0.71 & \bf 0.91 & \bf 0.66 & \bf 0.75 \\
				mBERT (cased)  & 0.61 & 0.67 & 0.64 & 0.84 & 0.77 & 0.54 & 0.89 & 0.69 & 0.71 &  & 0.58 & \bf 0.74 & 0.76 & \bf 0.81 & 0.77 & \bf 0.71 & \bf 0.91 & 0.54 & 0.73 \\
				mBERT (uncased)  & \bf 0.64 & 0.68 & 0.67 & 0.86 & 0.79 & 0.48 & 0.89 & 0.70 & 0.72 &  & 0.63 & 0.72 & 0.73 & \bf 0.81 & 0.77 & 0.69 & 0.90 & 0.57 & 0.73 \\
				XLM (base) & 0.62 & 0.65 & 0.59 & 0.86 & 0.80 & 0.39 & 0.87 & 0.63 & 0.68 &  & 0.61 & 0.64 & 0.69 & \bf 0.81 & 0.78 & 0.64 & 0.90 & 0.56 & 0.70 \\
				XLM (large) & 0.63 & 0.65 & 0.64 & 0.87 & 0.80 & 0.35 & 0.88 & 0.67 & 0.68 &  & \bf 0.64 & 0.66 & 0.69 & 0.80 & 0.77 & 0.66 & 0.90 & 0.57 & 0.71 \\
							
				\bottomrule
			\end{tabular}
		\end{adjustbox}
	\end{center}
	\caption{{Pearson} correlation ($r$) between automatic metrics and human judgments for \textbf{focus}. We compute the precision for ROUGE and BERTScore. BERTScore uses the optimized layer, and other metrics are computed by using default configuration of the original implementation.}
	\label{tab:focus_result}
\end{table*}

\begin{table*}[t!]
	\footnotesize
	\begin{center}
		\begin{adjustbox}{width=1\linewidth}
			\begin{tabular}{lrrrrrrrrrrrrrrrrrrr}
				\toprule
				\multirow{2}{*}{\textbf{Metrics}} & \multicolumn{9}{c}{\textbf{POINTER GENERATOR}} && \multicolumn{9}{c}{\textbf{BERT}}\\
				\cmidrule{2-10}
				\cmidrule{12-20}
				& \textbf{EN} & \textbf{ID} & \textbf{FR} & \textbf{TR} & \textbf{ZH} & \textbf{RU} & \textbf{DE} & \textbf{ES} & \textbf{Avg} & & \textbf{EN} & \textbf{ID} & \textbf{FR} & \textbf{TR} & \textbf{ZH} & \textbf{RU} & \textbf{DE} & \textbf{ES} & \textbf{Avg} \\
				\midrule
				\multicolumn{20}{l}{\textbf{Traditional Metrics}} \\
				\midrule
				ROUGE-1 & 0.59 & 0.73 & 0.66 & 0.79 & \bf 0.82 & 0.52 & 0.90 & 0.67 & 0.71 &  & 0.64 & 0.71 & 0.70 & 0.86 & 0.76 & 0.57 & 0.89 & 0.70 & 0.73 \\
				ROUGE-2 & 0.55 & 0.65 & 0.68 & 0.76 & 0.78 & 0.63 & 0.89 & 0.64 & 0.70 &  & 0.57 & 0.66 & 0.72 & 0.83 & 0.72 & 0.52 & 0.90 & 0.73 & 0.71 \\
				ROUGE-3 & 0.49 & 0.59 & 0.58 & 0.69 & 0.68 & 0.50 & 0.88 & 0.57 & 0.62 &  & 0.48 & 0.55 & 0.66 & 0.79 & 0.63 & 0.41 & 0.89 & 0.64 & 0.63 \\
				ROUGE-L & 0.58 & 0.74 & 0.64 & 0.79 & \bf 0.82 & 0.54 & 0.90 & 0.66 & 0.71 &  & 0.63 & 0.70 & 0.71 & 0.86 & 0.77 & 0.56 & 0.90 & 0.72 & 0.73 \\
				ROUGE-S & 0.60 & 0.69 & 0.63 & 0.74 & 0.77 & 0.52 & 0.89 & 0.66 & 0.69 &  & 0.61 & 0.69 & 0.69 & 0.81 & 0.70 & 0.51 & 0.89 & 0.71 & 0.70 \\
				ROUGE-SU & 0.59 & 0.70 & 0.64 & 0.75 & 0.79 & 0.52 & 0.89 & 0.67 & 0.69 &  & 0.61 & 0.69 & 0.70 & 0.82 & 0.71 & 0.57 & 0.89 & 0.71 & 0.71 \\
				ROUGE-W.12 & 0.54 & 0.71 & 0.64 & 0.77 & 0.81 & 0.55 & 0.90 & 0.65 & 0.69 &  & 0.61 & 0.68 & 0.69 & 0.85 & 0.75 & 0.56 & 0.90 & 0.68 & 0.71 \\
				METEOR & 0.60 & 0.72 & 0.65 & 0.77 & 0.81 & 0.55 & 0.89 & 0.63 & 0.70 &  & 0.66 & 0.69 & 0.69 & 0.83 & 0.75 & 0.59 & 0.89 & 0.75 & 0.73 \\
				BLEU-4 & 0.48 & 0.61 & 0.63 & 0.61 & 0.70 & 0.49 & 0.84 & 0.50 & 0.61 &  & 0.49 & 0.56 & 0.59 & 0.75 & 0.67 & 0.54 & 0.87 & 0.59 & 0.63 \\
				
				\midrule
				\multicolumn{20}{l}{\textbf{MoverScore}} \\
				\midrule
				mono-BERT & 0.57 & 0.65 & 0.63 & 0.73 & 0.79 & 0.68 & 0.86 & 0.55 & 0.68 &  & 0.61 & 0.60 & 0.69 & 0.86 & 0.75 & 0.66 & 0.91 & 0.68 & 0.72 \\
				mBERT (cased)  & 0.53 & 0.67 & 0.68 & 0.71 & 0.77 & 0.60 & 0.82 & 0.63 & 0.68 &  & 0.53 & 0.71 & 0.75 & 0.82 & 0.73 & 0.63 & 0.89 & 0.74 & 0.73 \\
				mBERT (uncased)  & 0.58 & 0.68 & 0.74 & 0.72 & 0.76 & 0.58 & 0.84 & 0.64 & 0.69 &  & 0.59 & 0.70 & 0.76 & 0.85 & 0.73 & 0.65 & 0.90 & 0.76 & 0.74 \\
				XLM (base) & 0.56 & 0.61 & 0.52 & 0.68 & 0.71 & 0.31 & 0.82 & 0.62 & 0.60 &  & 0.58 & 0.64 & 0.68 & 0.83 & 0.65 & 0.52 & 0.90 & 0.68 & 0.68 \\
				XLM (large) & 0.52 & 0.62 & 0.50 & 0.66 & 0.59 & 0.31 & 0.82 & 0.49 & 0.56 &  & 0.57 & 0.61 & 0.63 & 0.82 & 0.56 & 0.48 & 0.88 & 0.63 & 0.65 \\
				
				\midrule
				\multicolumn{20}{l}{\textbf{BERTScore}} \\
				\midrule
				mono-BERT & 0.63 & 0.74 & 0.76 & \bf 0.87 & 0.81 & \bf 0.72 & \bf 0.92 & \bf 0.73 & \bf 0.77 &  & 0.67 & \bf 0.74 & \bf 0.78 & \bf 0.89 & \bf 0.78 & 0.63 & \bf 0.92 & \bf 0.78 & \bf 0.77 \\
				mBERT (cased)  & \bf 0.67 & \bf 0.75 & 0.67 & 0.85 & \bf 0.82 & 0.70 & 0.91 & 0.72 & 0.76 &  & \bf 0.68 & 0.71 & 0.74 & \bf 0.89 & 0.76 & \bf 0.69 & 0.90 & 0.72 & 0.76 \\
				mBERT (uncased)  & 0.63 & \bf 0.75 & 0.70 & 0.85 & 0.81 & 0.67 & 0.91 & 0.71 & 0.76 &  & 0.64 & 0.73 & 0.76 & \bf 0.89 & 0.77 & 0.68 & 0.90 & 0.73 & 0.76 \\
				XLM (base) & 0.66 & 0.72 & 0.68 & 0.84 & 0.77 & 0.63 & 0.91 & 0.70 & 0.74 &  & 0.64 & 0.70 & 0.67 & 0.88 & 0.69 & 0.67 & 0.89 & 0.71 & 0.73 \\
				XLM (large) & 0.66 & 0.70 & 0.68 & 0.84 & 0.77 & 0.59 & 0.91 & 0.70 & 0.73 &  & 0.66 & 0.69 & 0.70 & 0.88 & 0.70 & \bf 0.69 & 0.90 & 0.72 & 0.74 \\

				\bottomrule
			\end{tabular}
		\end{adjustbox}
	\end{center}
	\caption{{Pearson} correlation ($r$) between automatic metrics and human judgments for \textbf{coverage}. We compute the recall for ROUGE and BERTScore. BERTScore uses the optimized layer, and other metrics are computed by using default configuration of the original implementation.}
	\label{tab:coverage_result}
\end{table*}

\begin{figure*}[h!]
	\centering
	\includegraphics[width=5in]{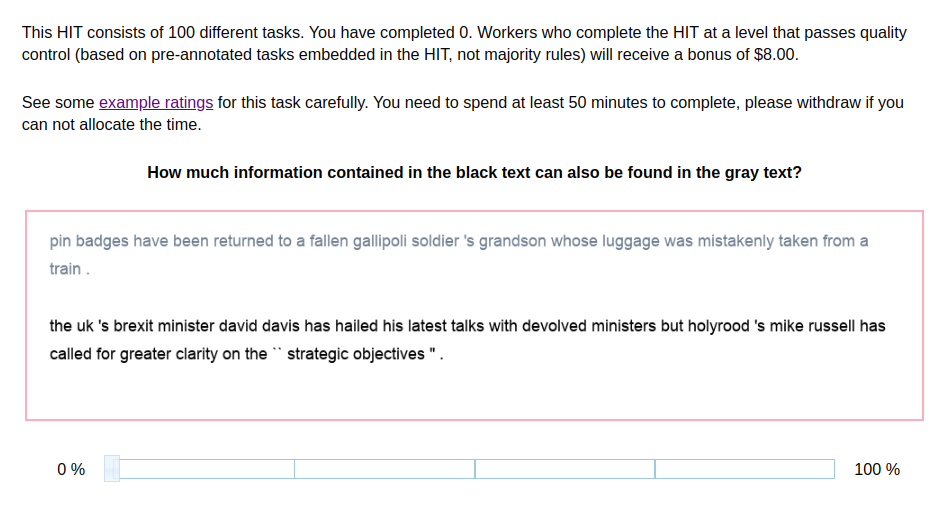}
	\caption{MTurk annotation interface for \textbf{English}.}
	\label{fig:interface_en}
\end{figure*}

\begin{figure*}[h!]
	\centering
	\includegraphics[width=5in]{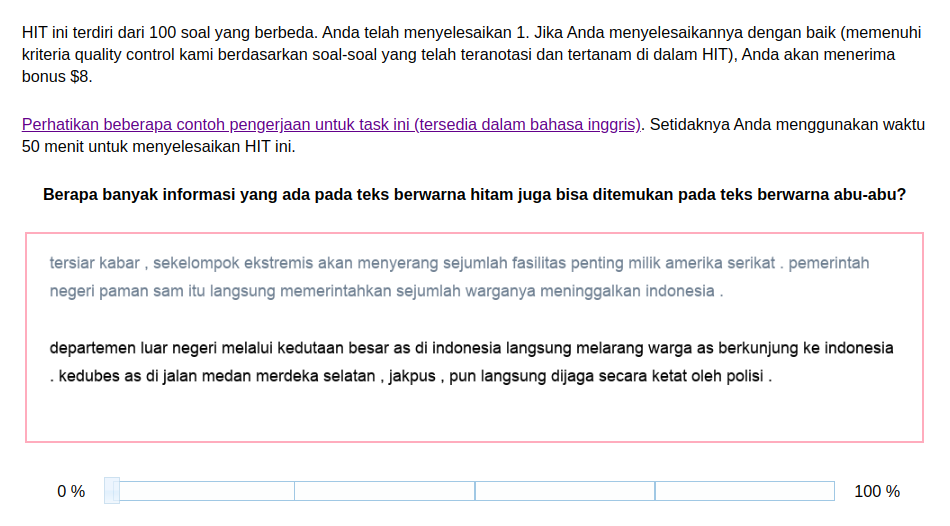}
	\caption{MTurk annotation interface for \textbf{Indonesian}.}
	\label{fig:interface_id}
\end{figure*}

\begin{figure*}[h!]
	\centering
	\includegraphics[width=5in]{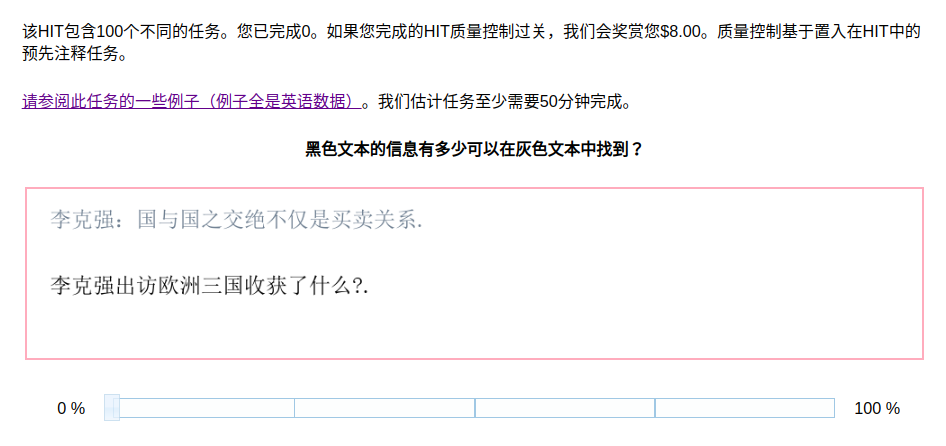}
	\caption{MTurk annotation interface for \textbf{Chinese}. Due to the page limit for the Appendix, the annotation interface for the other languages can be found at \url{https://github.com/fajri91/Multi_SummEval}}
	\label{fig:interface_zh}
\end{figure*}

%
%
%
%

\begin{table*}[t!]
	\footnotesize
	\begin{center}
		\begin{adjustbox}{width=1\linewidth}
			\begin{tabular}{lrrrrrrrrrrrrrrrrrrr}
				\toprule
				& \multicolumn{9}{c}{\textbf{Focus}} && \multicolumn{9}{c}{\textbf{Coverage}}\\
				\cmidrule{2-10}
				\cmidrule{12-20}
				& \textbf{EN} & \textbf{ID} & \textbf{FR} & \textbf{TR} & \textbf{ZH} & \textbf{RU} & \textbf{DE} & \textbf{ES} & \textbf{Avg} & & \textbf{EN} & \textbf{ID} & \textbf{FR} & \textbf{TR} & \textbf{ZH} & \textbf{RU} & \textbf{DE} & \textbf{ES} & \textbf{Avg} \\
				\midrule
				\multicolumn{20}{l}{\textbf{Traditional Metrics}} \\
				\midrule
				ROUGE-1 & 0.59 & 0.69 & 0.62 & 0.75 & 0.80 & 0.33 & 0.77 & 0.45 & 0.63 &  & 0.58 & 0.71 & 0.56 & 0.79 & 0.78 & 0.47 & 0.75 & 0.59 & 0.65 \\
				ROUGE-2 & 0.58 & 0.63 & 0.64 & 0.75 & 0.78 & 0.37 & 0.78 & 0.56 & 0.64 &  & 0.54 & 0.65 & 0.62 & 0.77 & 0.75 & 0.39 & 0.77 & 0.63 & 0.64 \\
				ROUGE-3 & 0.49 & 0.55 & 0.62 & 0.70 & 0.68 & 0.27 & 0.77 & 0.51 & 0.58 &  & 0.47 & 0.58 & 0.59 & 0.69 & 0.65 & 0.28 & 0.75 & 0.58 & 0.57 \\
				ROUGE-L & 0.60 & 0.69 & 0.62 & 0.75 & 0.80 & 0.32 & 0.76 & 0.46 & 0.62 &  & 0.57 & 0.70 & 0.55 & 0.79 & 0.78 & 0.47 & 0.74 & 0.58 & 0.65 \\
				ROUGE-S & 0.61 & 0.71 & 0.61 & 0.75 & 0.81 & 0.36 & 0.77 & 0.45 & 0.63 &  & 0.60 & 0.72 & 0.56 & 0.79 & 0.79 & 0.39 & 0.75 & 0.58 & 0.65 \\
				ROUGE-SU & 0.61 & 0.71 & 0.61 & 0.74 & 0.81 & 0.31 & 0.77 & 0.43 & 0.62 &  & 0.59 & 0.72 & 0.56 & 0.79 & 0.79 & 0.45 & 0.75 & 0.58 & 0.66 \\
				ROUGE-W.12 & 0.61 & 0.67 & 0.61 & 0.74 & 0.79 & 0.32 & 0.76 & 0.47 & 0.62 &  & 0.55 & 0.68 & 0.55 & 0.78 & 0.78 & 0.48 & 0.74 & 0.58 & 0.64 \\
				METEOR & 0.47 & 0.68 & 0.60 & 0.74 & 0.82 & 0.40 & 0.75 & 0.54 & 0.62 &  & 0.61 & 0.70 & 0.59 & 0.81 & 0.78 & 0.44 & 0.74 & 0.62 & 0.66 \\
				BLEU-4 & 0.50 & 0.59 & 0.62 & 0.67 & 0.76 & 0.05 & 0.77 & 0.51 & 0.56 &  & 0.49 & 0.61 & 0.58 & 0.62 & 0.72 & -0.04 & 0.73 & 0.55 & 0.53 \\
								
				\midrule
				\multicolumn{20}{l}{\textbf{MoverScore}} \\
				\midrule
				mono-BERT & 0.58 & 0.65 & 0.66 & \bf 0.81 & 0.79 & 0.38 & \bf 0.85 & 0.55 & 0.66 &  & 0.59 & 0.61 & 0.60 & 0.81 & 0.77 & 0.30 & \bf 0.83 & 0.59 & 0.64 \\
				mBERT (cased)  & 0.53 & 0.68 & 0.73 & 0.80 & 0.77 & 0.44 & 0.83 & 0.60 & 0.67 &  & 0.50 & 0.68 & 0.65 & 0.77 & 0.74 & 0.37 & 0.79 & 0.65 & 0.64 \\
				mBERT (uncased)  & \bf 0.60 & 0.69 & \bf 0.74 & 0.80 & 0.77 & \bf 0.45 & 0.83 & \bf 0.66 & \bf 0.69 &  & 0.58 & 0.68 & 0.68 & 0.77 & 0.74 & 0.42 & 0.79 & 0.67 & 0.67 \\
				XLM (base) & 0.51 & 0.63 & 0.64 & 0.78 & 0.70 & 0.11 & 0.78 & 0.53 & 0.59 &  & 0.55 & 0.62 & 0.58 & 0.73 & 0.65 & 0.01 & 0.75 & 0.59 & 0.56 \\
				XLM (large) & 0.51 & 0.59 & 0.63 & 0.72 & 0.54 & 0.10 & 0.76 & 0.44 & 0.54 &  & 0.54 & 0.61 & 0.52 & 0.68 & 0.52 & 0.07 & 0.72 & 0.50 & 0.52 \\
								
				\midrule
				\multicolumn{20}{l}{\textbf{BERTScore}} \\
				\midrule
				mono-BERT & 0.58 & 0.70 & 0.67 & 0.77 & \bf 0.81 & 0.34 & 0.84 & 0.60 & 0.66 &  & \bf 0.64 & \bf 0.73 & \bf 0.70 & \bf 0.86 & \bf 0.78 & 0.57 & \bf 0.83 & \bf 0.70 & \bf 0.73 \\
				mBERT (cased)  & 0.53 & \bf 0.71 & 0.68 & 0.78 & 0.78 & 0.41 & 0.82 & 0.55 & 0.66 &  & \bf 0.64 & 0.72 & 0.58 & 0.85 & 0.77 & \bf 0.63 & 0.77 & 0.67 & 0.70 \\
				mBERT (uncased)  & 0.58 & 0.70 & 0.66 & 0.79 & 0.79 & 0.40 & 0.83 & 0.58 & 0.67 &  & 0.60 & 0.73 & 0.60 & 0.85 & 0.77 & \bf 0.63 & 0.77 & 0.66 & 0.70 \\
				XLM (base) & 0.56 & 0.65 & 0.61 & 0.78 & 0.78 & 0.04 & 0.78 & 0.49 & 0.59 &  & 0.62 & 0.71 & 0.58 & 0.84 & 0.70 & 0.55 & 0.76 & 0.64 & 0.67 \\
				XLM (large) & 0.57 & 0.66 & 0.62 & 0.78 & 0.77 & 0.17 & 0.80 & 0.52 & 0.61 &  & 0.62 & 0.69 & 0.60 & 0.84 & 0.72 & 0.50 & 0.78 & 0.65 & 0.68 \\
				\bottomrule
			\end{tabular}
		\end{adjustbox}
	\end{center}
	\caption{{Spearman} correlation ($\rho$) between automatic metrics and human 
		judgments (for Pointer Generator and BERT models combined). We compute the precision and recall of ROUGE and BERTScore for focus and coverage, respectively. BERTScore uses the optimized layer, and other metrics are computed by using default configuration of the original implementation.}
	\label{tab:result_spearmanr}
\end{table*}

\end{document}